\documentclass[a4paper]{article}

\usepackage{INTERSPEECH2022}
\usepackage{multirow}
\usepackage{caption}
\usepackage{subcaption}
\usepackage{enumitem}
\usepackage{amsmath}
\DeclareMathOperator*{\argmax}{arg\,max}

\usepackage{caption}
\usepackage{subcaption}
\usepackage{enumitem}
\usepackage{hyperref}

\title{Does Utterance entails Intent?: Evaluating Natural Language Inference
Based Setup for Few-Shot Intent Detection}
\name{Author Name$^1$, Co-author Name$^2$}
\name{Ayush Kumar$^{\dagger}$, Vijit Malik$^{\dagger\star}$\thanks{$^{\dagger}$Equal contribution. $^{\star}$Work done during internship at Observe.AI}, Jithendra Vepa}
\address{Observe.AI, India}
\email{ ayush@observe.ai, vijitvm@iitk.ac.in, jithendra@observe.ai}

\begin{document}

\maketitle
\begin{abstract}
Intent Detection is one of the core tasks of 
dialog systems. Few-shot Intent Detection is challenging due
to limited number of annotated utterances for novel classes. Generalized Few-shot intent detection is more realistic but challenging setup which aims to discriminate the joint label space of both novel intents which have few examples each and existing intents consisting of enough labeled data. Large label spaces and fewer number of shots increase the complexity of the task. In this work, we employ a simple and effective method based on Natural Language Inference that leverages the semantics in the class-label names to learn and predict the novel classes. Our method achieves state-of-the-art results on 1-shot and 5-shot intent detection task with gains ranging from 2-8\% points in F1 score on four benchmark datasets. 
Our method also outperforms existing approaches on
a more practical setting of generalized few-shot intent detection with gains up to 20\% F1 score. 
We show that the suggested approach performs well across single and multi domain datasets with the number of class labels from as few as 7 to as high as 150.
\end{abstract}

\section{Introduction}
Intent Detection (ID) task aims to identify the user's goals behind their utterances. 
Detecting the intent in these utterances is a crucial step in spoken language understanding systems.
Supervised approaches for the task require a large number of labelled samples for each intent class to train upon. This requirement inhibits the model's ability to generalize to emerging (novel) intents with no or limited annotations.
 
Few-shot Learning (FSL) is a paradigm adopted to address the challenges of emerging (novel) classes. The task is to discriminate the novel classes from each other given only a few training samples of each class \cite{DBLP:conf/cvpr/MillerMV00}.
This small set of labeled examples for novel classes is  referred to as \textit{support set}.
In addition to a support set, we also have access to a \textit{seen set} containing sufficient examples of seen classes.
In a more pragmatic setting of SLU systems, the new intents emerge over the time once a model has been trained on seen classes. Thus, the model is expected to discriminate between the classes in a joint label space of seen and novel classes during inference. This task is referred as a \textit{generalized few-shot intent detection} (GFSID) task \cite{DBLP:journals/corr/abs-2004-01881}.

For the task of few-shot intent detection (FSID), a set of works utilize prototypical networks \cite{DBLP:conf/nips/SnellSZ17, xu2021semantic, sun2019hierarchical} that creates a representative vector for each class and use various distance metrics 
to infer the class of a new data sample. 
Other works propose multi-level matching and aggregation methods to improve FSL performance \cite{nguyen-etal-2020-dynamic, ye-ling-2019-multi}.
The primary idea in matching based methods is to learn associations between multiple semantic components from support and query samples. We note that even with multi-level matching and attention networks, these methods do not perform well in GFSID and large label space settings. 
A work in computer vision use  Model-Agnostic Meta-Learning (MAML) approach \cite{finn2017model} from multiple subtasks.
Compared to metric based (prototypical, matching) approaches, Li et al. \cite{DBLP:conf/slt/0001KDZA21} find that optimization based (MAML) method perform poorly on intent detection task. Therefore, in this work we focus our comparison with metric-based approaches. 
A set of works on FSID utilize approaches 
to generate samples for the support set and convert FSID to a fully-supervised text classification problem \cite{DBLP:journals/corr/abs-2004-01881, xia-etal-2020-composed, 10.1145/3404835.3462995}. 
However, an inherent assumption of representing the intents as a pair of action-object tokens present in the utterances limits the generalizability of these methods to all datapoints. For example, an utterance \textit{`need a place for italian dinner'} corresponding to the `book restaurant' intent does not contain action-token related to 'book' or object-token related to 'restaurant'. Additionally, such generative methods suffer from propagation of errors in generated texts while training the FSID model on the generated samples.

In this work, we draw inspiration from Natural Language Inference (NLI) \cite{bowman-etal-2015-large} and
convert the intent detection task into a textual entailment task that measures the truth value of the hypothesis (\textit{class-label}) given a premise (\textit{utterance}). 
Prior work  
shows that NLI can be used as a unified language processing method \cite{DBLP:conf/emnlp/YinRRSX20}. Sainz et al., \cite{sainz-etal-2021-label} attempts to utilize NLI based few-shot learning for Relation Extraction whereas Zhang et al. \cite{zhang-etal-2020-discriminative} focuses upon the detection of out-of-scope intents in a multi domain dataset.
The efficacy of using natural language inference based approach is not yet studied for both few-shot and generalized few-shot intent detection across varying size of seen and unseen class labels.
In this work, we put forth a focused approach that quantifies the performance of NLI based approach for the task of few-shot (and generalized) intent detection.
In summary, our contributions are three-fold:
\begin{itemize}
    \item We employ a simple yet highly effective approach based on NLI for few-shot and generalized few-shot intent detection. Our method is a two-step process that involves a data transformation for entailment as the initial step and fine-tuning the base BERT model on the entailment task as the final step.
    \item We carry out experiments to demonstrate the efficacy of the approach on multiple benchmark intent datasets.
    We show that our method is distinctly effective for large label space and GFSID settings. 
    \item 
    To help advance the exploration and research on few-shot intent detection, we release the \href{https://github.com/Observeai-Research/NLI-FSL} {codebase} developed for this work.
\end{itemize}
\section{Problem Statement}
The task of intent detection is to determine the intent of the utterance $x$ given the set of intent classes, $\mathcal{Y}$. 
The goal of few-shot classification is to learn a classifier $f_{\phi}$ for a set of $n$ novel classes $\mathcal{Y}_{n} = \{C_{1}, C_{2},... C_{n}\}$, where $C_i$ denotes the class-label name of intent $i$. In a traditional FSL task, the classifier performs a series of tasks during both training and inference, which involves $\mathcal{C}$ randomly chosen classes with only $\mathcal{K}$ labeled samples from each class. This is called a $\mathcal{C}$-way $\mathcal{K}$-shot classification task. The set of $\mathcal{C.K}$ samples is called the support set. 
This series of tasks are repeated via episodes \cite{vinyals2016matching} where in each episode, the aim is to correctly classify unlabeled samples (query samples)
by using only the support samples. 
In an episodic setting, training is organized in a series of meta learning problems (or episodes), each divided into a small training and validation subset to mimic the circumstances encountered during evaluation. However, episodic setting 
does not provide an end-to-end systematic evaluation in practical applications \cite{laenen2021on}, where the requirement is to categorize unlabeled samples into one of the novel/joint classes instead of a set of sampled classes. Thus, we evaluate our method using the non-episodic setting.

In addition to the set of novel classes $\mathcal{Y}_{n}$, we also have a set of seen classes $\mathcal{Y}_{s} = \{C'_{1}, C'_{2},...C'_{s}\}$. Note that the sets $\mathcal{Y}_{s}$ and $\mathcal{Y}_{n}$ are disjoint. We denote seen-class dataset as the set of examples $\mathcal{D}_{s}$ $= \{(x_{1},y_{1}), ... (x_{|\mathcal{D}_{s}|},y_{|\mathcal{D}_{s}|})\}$, where $y_{i} \in \mathcal{Y}_{s}$. Similarly, the novel-class (support) dataset is denoted as $\mathcal{D}_{n}$ with label as $\mathcal{Y}_{n}$.
In few-shot intent detection (FSID), the task is to maximize the probability of the correct prediction for an unlabeled utterance $u$ in the novel label subset, $\mathcal{Y}_{n}$ as

\begin{equation}
    \hat{y} = \argmax_{y \in \mathcal{Y}_{n}}p(y|u,\mathcal{D}_{n},\mathcal{D}_{s})
\end{equation}

In generalized few-shot intent detection (GFSID), the task is to maximize the probability of the correct prediction for an unlabeled utterance $u$ in the joint label subspace, $\mathcal{Y}_{j}$ (where $\mathcal{Y}_{j} = \mathcal{Y}_{n} \cup \mathcal{Y}_{s}$) as

\begin{equation}
    \hat{y} = \argmax_{y \in \mathcal{Y}_{j}}p(y|u,\mathcal{D}_{n}, \mathcal{D}_{s})
\end{equation}
\section{Methodology}

\begin{figure}[ht]
     \centering
    \includegraphics[width=0.95\linewidth,  height=8.5cm]{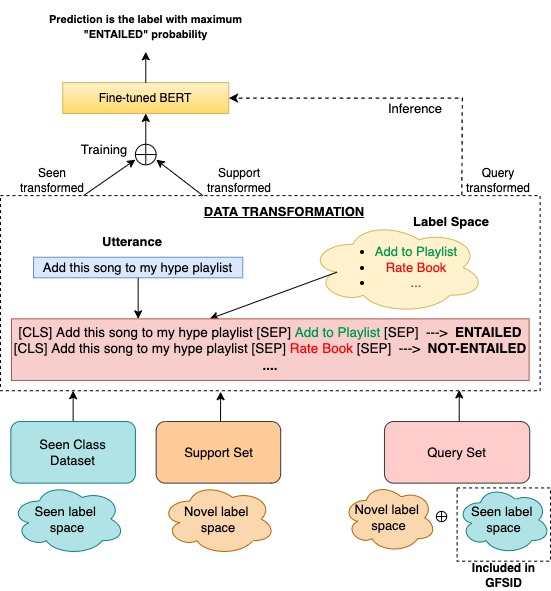}
         \caption{NLI-FSL methodology}
         \label{fig:method}
\end{figure}

Natural language inference is defined as the task of determining whether or not a premise semantically entails a hypothesis. The generality of the task format makes it a unified method to model the classification tasks. We conjecture that intent's class-label names such as \textit{add\_to\_playlist, rate\_book etc.} carry semantics that reflects the meaning of utterances in respective intents. Thus, we transform the task of intent detection to textual entailment by simultaneously setting the input utterance as the premise and casting the intent's class-label name as the hypothesis.

Our proposed approach named \textit{NLI-FSL} is a two step process where we first transform the raw dataset into a NLI-formatted dataset using samples of both positive (entailed) and negative (not-entailed) examples. In the next step, we fine-tune a pre-trained language model (PLM) upon the transformed dataset and then use it for inference (Figure \ref{fig:method}). We utilize the BERT \cite{devlin-etal-2019-bert} model as the PLM in all our experiments. 

To frame an NLI-like task for FSID, we make the model learn a pair of utterance and it's correct class-label name (positive pair) to represent entailment while a pair of utterance and a randomly selected class-label name (negative pair) to denote a scenario of contradiction/non-entailment. For an utterance $u$ 
we make $|\mathcal{Y}|$ inputs corresponding to each label $y \in \mathcal{Y}$ in the form of \texttt{[CLS]$w_{1} w_{2}... w_{N_{u}}$[SEP]$l_{1} l_{2}...l_{N_{y}}$[SEP]} where $l_{1} l_{2}...l_{N_{y}}$ denotes the tokenized label-name $y$. 
The output space consists of only two labels, $0$ for not-entailed and $1$ for entailed. Note that for each utterance $u$, we end up with exactly $1$ entailed pair $(u,y_{t})$ and $(|\mathcal{Y}|-1)$ not-entailed pairs $(u,y)$, where $y_t$ denotes the correct class-label for utterance $u$ and $y\in\{\mathcal{Y}\symbol{92}\{y_{t}\}\}$. 

In the first step of our methodology, we transform the seen set $\mathcal{D}_{s}$ to the NLI format, $\mathcal{D'}_{s}$ with a total of $|\mathcal{D}_{s}|*(|\mathcal{Y}_{s}|-1)$ non-entailment and $|\mathcal{D}_{s}|$ entailment pairs. Similarly, we take the support set $\mathcal{D}_{n}$ and obtain the transformed set $\mathcal{D'}_{n}$. 
Due to the heavy class imbalance between entailment and non-entailment classes, we randomly sample some instances from  $\mathcal{D'}_{s}$ and $\mathcal{D'}_{n}$ having the non-entailment label for further use. Empirically, we find the ratio of 2:1 between non-entailment to entailment pairs work best in our experiments.
After creating NLI-format dataset, in second step, we fine-tune the BERT-base model on the combination of sampled $\mathcal{D'}_{s}$ and $\mathcal{D'}_{n}$ to learn discriminating between entailment and non-entailment pairs. 

During inference, we transform the query/test set into the NLI form, similar to the training dataset. For each test utterance $u$ and the label space $\mathcal{Y}$, the predicted label is determined by the intent having highest probability for entailment label $e$: 
\begin{equation}
    y_{pred} = \argmax_{y_{i} \in \mathcal{Y}}p(e|y_{i},u)
\end{equation}

\section{Datasets}

We evaluate our proposed methodology over five benchmark Intent Detection (ID) datasets for FSID/GFSID tasks. These datasets represent real world queries in different domains such as banking, airline and customer-service described as:
\textit{a)} \textbf{SNIPS} is a dataset of crowd-sourced queries distributed among 7 user intents of diverse complexity \cite{Coucke2018SnipsVP}. We use three intents in the novel set and four in the seen set;
\textit{b)} Airline Travel Information System (\textbf{ATIS}) is a domain specific intent detection benchmark for flight reservation queries \cite{hemphill-etal-1990-atis}. In line with previous works, we select 4 classes as the novel classes and remaining 12 as seen class;
\textit{c)} \textbf{BANKING77} is a single domain dataset composed of online banking queries annotated with their corresponding intents \cite{DBLP:journals/corr/abs-2106-04564}. We use 27 classes from 77 classes as novel and take the remaining 50 classes as seen classes;
\textit{d)} In order to compare the method on a  dataset containing higher number of labels, we utilize \textbf{CLINC150} dataset \cite{larson-etal-2019-evaluation}. It is a multi domain intent detection dataset containing a total of 150 classes. We take 50 classes 
as novel and take the remaining 100 classes as seen classes. Similar to previous works, we randomly choose seen and novel classes wherever the split is not available.

\section{Baselines}
We compare our methodology with several competitive works: 
\begin{itemize}

    \item \textbf{Prototypical Network with BERT-base encoder (ProtoBERT)} \cite{DBLP:conf/nips/SnellSZ17} fine-tuned the BERT-base model upon the seen class dataset for supervised Intent Detection and uses the fine-tuned BERT model to obtain class prototypes and calculates Euclidean distance from query and prototypes for classification.
    \item \textbf{Multi-level Matching and Aggregation Network (MLMAN)} \cite{ye-ling-2019-multi}  utilized the multi-level matching approach and exploits both fusion and dot product similarity for better instance representation.
    \item \textbf{Hybrid Attention-Based Prototypical Network (HATT)} \cite{gao2019hybrid} used instance level and feature-level attention schemes based on prototypical networks to highlight the crucial instances and features.
    \item \textbf{Semantic Aggregation and Matching (SMAN)} \cite{nguyen-etal-2020-dynamic} distilled the semantic components are from utterances via multi-head self-attention by imposing additional dynamic regularization constraints for few-shot learning.
    \item \textbf{Discriminative Nearest Neighbour Classification (DNNC)} \cite{zhang-etal-2020-discriminative} is a state-of-art method that leverages nearest-neighbor classification model to detect few-shot user intents and OOS intents by boosting the ability of the model to discriminate between the classes.
\end{itemize}


\section{Experiments and Results}

\begin{table*}[ht]
\caption{Comparison of results by different methods. All results are reported in macro-F1. \textsuperscript{\textdagger} indicate that results are significant improvement ($p < 0.05$) over baselines under t-test.}
\small
\renewcommand{\arraystretch}{1.1}
\centering
\begin{tabular}{|cccclccc|}
\hline
\multicolumn{1}{|c|}{\multirow{2}{*}{\textbf{Method}}} & \multicolumn{3}{c|}{\textbf{1-shot}}                                                                               & \multicolumn{1}{l|}{} & \multicolumn{3}{c|}{\textbf{5-shot}}                                                          \\ \cline{2-4} \cline{6-8} 
\multicolumn{1}{|c|}{}                                 & \multicolumn{1}{c|}{\textbf{FSID}}  & \multicolumn{1}{c|}{\textbf{GFSID}} & \multicolumn{1}{c|}{\textbf{Harmonic}} & \multicolumn{1}{l|}{} & \multicolumn{1}{c|}{\textbf{FSID}}  & \multicolumn{1}{c|}{\textbf{GFSID}} & \textbf{Harmonic} \\ \hline
\multicolumn{8}{|c|}{\textbf{SNIPS}}                                                                                                                                                                                                                                                                \\ \hline
\multicolumn{1}{|c|}{ProtoBERT}                        & \multicolumn{1}{c|}{66.58}          & \multicolumn{1}{c|}{59.32}          & \multicolumn{1}{c|}{62.74}             & \multicolumn{1}{l|}{} & \multicolumn{1}{c|}{83.77}          & \multicolumn{1}{c|}{78.39}          & 80.99             \\ \cline{1-4} \cline{6-8} 
\multicolumn{1}{|c|}{MLMAN}                            & \multicolumn{1}{c|}{68.22}          & \multicolumn{1}{c|}{62.95}          & \multicolumn{1}{c|}{65.48}             & \multicolumn{1}{l|}{} & \multicolumn{1}{c|}{85.74}          & \multicolumn{1}{c|}{79.42}          & 82.46             \\ \cline{1-4} \cline{6-8} 
\multicolumn{1}{|c|}{HATT}                             & \multicolumn{1}{c|}{67.97}          & \multicolumn{1}{c|}{64.04}          & \multicolumn{1}{c|}{65.95}             & \multicolumn{1}{l|}{} & \multicolumn{1}{c|}{87.29}          & \multicolumn{1}{c|}{81.57}        & 84.33             \\ \cline{1-4} \cline{6-8} 
\multicolumn{1}{|c|}{SMAN}                             & \multicolumn{1}{c|}{70.58}          & \multicolumn{1}{c|}{68.85}          & \multicolumn{1}{c|}{69.70}             & \multicolumn{1}{l|}{} & \multicolumn{1}{c|}{87.80}          & \multicolumn{1}{c|}{83.05}          & 85.36             \\ \cline{1-4} \cline{6-8} 
\multicolumn{1}{|c|}{DNNC}                             & \multicolumn{1}{c|}{78.48}          & \multicolumn{1}{c|}{67.86}          & \multicolumn{1}{c|}{72.78}             & \multicolumn{1}{l|}{} & \multicolumn{1}{c|}{90.31}          & \multicolumn{1}{c|}{83.64}          & 86.85             \\ \cline{1-4} \cline{6-8} 
\multicolumn{1}{|c|}{NLI-FSL}                          & \multicolumn{1}{c|}{\textbf{86.40\textsuperscript{\textdagger}}} & \multicolumn{1}{c|}{\textbf{74.32\textsuperscript{\textdagger}}} & \multicolumn{1}{c|}{\textbf{79.91}}    & \multicolumn{1}{l|}{} & \multicolumn{1}{c|}{\textbf{92.07\textsuperscript{\textdagger}}} & \multicolumn{1}{c|}{\textbf{88.91\textsuperscript{\textdagger}}} & \textbf{90.46}    \\ \hline

\multicolumn{8}{|c|}{\textbf{ATIS}}                                                                                                                                                                                                                                                                 \\ \hline
\multicolumn{1}{|c|}{ProtoBERT}                        & \multicolumn{1}{c|}{74.14}          & \multicolumn{1}{c|}{26.88}          & \multicolumn{1}{c|}{39.46}             & \multicolumn{1}{l|}{} & \multicolumn{1}{c|}{73.28}          & \multicolumn{1}{c|}{41.09}          & 52.65             \\ \cline{1-4} \cline{6-8} 
\multicolumn{1}{|c|}{MLMAN}                            & \multicolumn{1}{c|}{73.94}          & \multicolumn{1}{c|}{35.27}          & \multicolumn{1}{c|}{47.78}             & \multicolumn{1}{l|}{} & \multicolumn{1}{c|}{78.24}          & \multicolumn{1}{c|}{47.33}          & 58.98             \\ \cline{1-4} \cline{6-8} 
\multicolumn{1}{|c|}{HATT}                             & \multicolumn{1}{c|}{74.03}          & \multicolumn{1}{c|}{39.39}          & \multicolumn{1}{c|}{51.89}             & \multicolumn{1}{l|}{} & \multicolumn{1}{c|}{80.62}          & \multicolumn{1}{c|}{50.24}          & 61.90             \\ \cline{1-4} \cline{6-8} 
\multicolumn{1}{|c|}{SMAN}                             & \multicolumn{1}{c|}{75.25}          & \multicolumn{1}{c|}{41.86}          & \multicolumn{1}{c|}{51.42}             & \multicolumn{1}{l|}{} & \multicolumn{1}{c|}{81.16}          & \multicolumn{1}{c|}{56.71}          & 66.77             \\ \cline{1-4} \cline{6-8} 
\multicolumn{1}{|c|}{DNNC}                             & \multicolumn{1}{c|}{74.38}          & \multicolumn{1}{c|}{44.91}          & \multicolumn{1}{c|}{56.05}             & \multicolumn{1}{l|}{} & \multicolumn{1}{c|}{80.98}          & \multicolumn{1}{c|}{59.36}          & 68.50             \\ \cline{1-4} \cline{6-8} 
\multicolumn{1}{|c|}{NLI-FSL}                          & \multicolumn{1}{c|}{\textbf{79.48\textsuperscript{\textdagger}}} & \multicolumn{1}{c|}{\textbf{55.71\textsuperscript{\textdagger}}} & \multicolumn{1}{c|}{\textbf{65.51}}    & \multicolumn{1}{l|}{} & \multicolumn{1}{c|}{\textbf{85.33\textsuperscript{\textdagger}}} & \multicolumn{1}{c|}{\textbf{66.80\textsuperscript{\textdagger}}} & \textbf{74.94}    \\ \hline

\multicolumn{8}{|c|}{\textbf{BANKING77}}                                                                                                                                                                                                                                                            \\ \hline
\multicolumn{1}{|c|}{ProtoBERT}                        & \multicolumn{1}{c|}{27.48}          & \multicolumn{1}{c|}{18.92}          & \multicolumn{1}{c|}{22.41}             & \multicolumn{1}{l|}{} & \multicolumn{1}{c|}{45.30}          & \multicolumn{1}{c|}{33.29}          & 38.38             \\ \cline{1-4} \cline{6-8} 
\multicolumn{1}{|c|}{MLMAN}                            & \multicolumn{1}{c|}{35.62}          & \multicolumn{1}{c|}{24.81}          & \multicolumn{1}{c|}{29.25}             & \multicolumn{1}{l|}{} & \multicolumn{1}{c|}{48.51}          & \multicolumn{1}{c|}{37.71}          & 42.43             \\ \cline{1-4} \cline{6-8} 
\multicolumn{1}{|c|}{HATT}                             & \multicolumn{1}{c|}{36.57}          & \multicolumn{1}{c|}{26.27}          & \multicolumn{1}{c|}{30.58}             & \multicolumn{1}{l|}{} & \multicolumn{1}{c|}{50.11}          & \multicolumn{1}{c|}{39.54}          & 44.20             \\ \cline{1-4} \cline{6-8} 
\multicolumn{1}{|c|}{SMAN}                             & \multicolumn{1}{c|}{42.94}          & \multicolumn{1}{c|}{27.40}          & \multicolumn{1}{c|}{33.45}             & \multicolumn{1}{l|}{} & \multicolumn{1}{c|}{52.30}          & \multicolumn{1}{c|}{40.75}          & 45.81             \\ \cline{1-4} \cline{6-8} 
\multicolumn{1}{|c|}{DNNC}                             & \multicolumn{1}{c|}{69.46}          & \multicolumn{1}{c|}{40.73}          & \multicolumn{1}{c|}{51.35}             & \multicolumn{1}{l|}{} & \multicolumn{1}{c|}{72.16}          & \multicolumn{1}{c|}{58.37}          & 64.54             \\ \cline{1-4} \cline{6-8} 
\multicolumn{1}{|c|}{NLI-FSL}                          & \multicolumn{1}{c|}{\textbf{74.31\textsuperscript{\textdagger}}} & \multicolumn{1}{c|}{\textbf{58.64\textsuperscript{\textdagger}}} & \multicolumn{1}{c|}{\textbf{65.55}}    & \multicolumn{1}{l|}{} & \multicolumn{1}{c|}{\textbf{79.79\textsuperscript{\textdagger}}} & \multicolumn{1}{c|}{\textbf{65.29\textsuperscript{\textdagger}}} & \textbf{71.82}    \\ \hline

\multicolumn{8}{|c|}{\textbf{CLINC150}}                                                                                                                                                                                                                                                             \\ \hline
\multicolumn{1}{|c|}{ProtoBERT}                        & \multicolumn{1}{c|}{29.76}          & \multicolumn{1}{c|}{23.71}          & \multicolumn{1}{c|}{26.39}             & \multicolumn{1}{l|}{} & \multicolumn{1}{c|}{61.31}          & \multicolumn{1}{c|}{50.20}          & 55.20             \\ \cline{1-4} \cline{6-8} 
\multicolumn{1}{|c|}{MLMAN}                            & \multicolumn{1}{c|}{34.29}          & \multicolumn{1}{c|}{25.66}          & \multicolumn{1}{c|}{29.35}             & \multicolumn{1}{l|}{} & \multicolumn{1}{c|}{63.79}          & \multicolumn{1}{c|}{54.55}          & 58.81             \\ \cline{1-4} \cline{6-8} 
\multicolumn{1}{|c|}{HATT}                             & \multicolumn{1}{c|}{36.57}          & \multicolumn{1}{c|}{28.04}          & \multicolumn{1}{c|}{31.74}             & \multicolumn{1}{l|}{} & \multicolumn{1}{c|}{65.23}          & \multicolumn{1}{c|}{56.93}          & 60.80             \\ \cline{1-4} \cline{6-8} 
\multicolumn{1}{|c|}{SMAN}                             & \multicolumn{1}{c|}{39.92}          & \multicolumn{1}{c|}{31.12}          & \multicolumn{1}{c|}{34.97}             & \multicolumn{1}{l|}{} & \multicolumn{1}{c|}{68.13}          & \multicolumn{1}{c|}{60.09}          & 63.86             \\ \cline{1-4} \cline{6-8} 
\multicolumn{1}{|c|}{DNNC}                             & \multicolumn{1}{c|}{71.35}          & \multicolumn{1}{c|}{53.47}          & \multicolumn{1}{c|}{61.13}             & \multicolumn{1}{l|}{} & \multicolumn{1}{c|}{80.15}          & \multicolumn{1}{c|}{65.73}          & 72.23             \\ \cline{1-4} \cline{6-8} 
\multicolumn{1}{|c|}{NLI-FSL}                          & \multicolumn{1}{c|}{\textbf{79.20\textsuperscript{\textdagger}}} & \multicolumn{1}{c|}{\textbf{74.02\textsuperscript{\textdagger}}} & \multicolumn{1}{c|}{\textbf{76.52}}    & \multicolumn{1}{l|}{} & \multicolumn{1}{c|}{\textbf{86.55\textsuperscript{\textdagger}}} & \multicolumn{1}{c|}{\textbf{73.49\textsuperscript{\textdagger}}} & \textbf{79.49}    \\ \hline

\end{tabular}
\label{tab:main_experiments}
\end{table*}

\subsection{Implementation details} 
We use the \texttt{BERT-base-uncased} model from the Huggingface\footnote{https://huggingface.co/} library. We fine-tune the model on the NLI-format datasets for 3 epochs for SNIPS and ATIS datasets and 4 epochs for BANKING77, NLUED, and CLINC150 datasets. To handle the class imbalance of \textit{entailed} and \textit{not-entailed} examples, we downsample the \textit{not-entailed} examples to be in the $r:1$ ratio with the number of \textit{entailed} examples, where $1<r<5$. The value of $r=2$ performs best in our experiments. The model is trained with a batch size of 64, learning rate of $2e$\text{-}$5$ and a binary cross-entropy loss using a AdamW optimizer with 1000 warmup steps on a K80 GPU. 
We use macro-F1 score as the evaluation metric for the experiments.
To measure the generalization to the FSID and GFSID settings, we also report the harmonic mean of the performances in each setting.
The evaluation for the baseline methods is carried with the publicly released codes of each baseline. We report the average results over five runs in each experiment setup. 


\subsection{Results and Discussion}

\begin{figure}[t]
     \centering
    \includegraphics[width=0.9\linewidth]{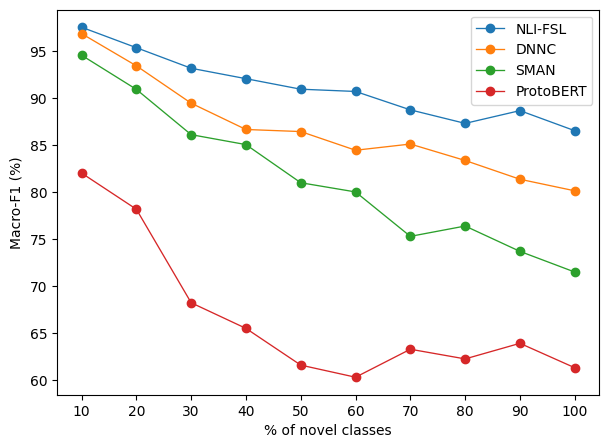}
    \caption{Incremental label space experiments on CLINC150.}
    \label{fig:inc_labels}
\end{figure}

We report the comparisons with all baselines on FSID and GFSID settings in Table \ref{tab:main_experiments}. The results show that NLI-FSL approach outperforms all methods in both 1-shot and 5-shot FSID and GFSID settings. Our method also produces best harmonic F1-score suggesting the effectiveness of the method in balancing its performance between both FSID and GFSID settings. 

Based on results, ProtoBERT comes out as the weakest baseline while DNNC is the strongest baseline. NLI-FSL outperforms DNNC by a margin up to 8\% macro-F1 in 1-shot and 5-shot FSID (Table \ref{tab:main_experiments}). A stronger performance by NLI-FSL is observed in GFSID settings, where the proposed method outperforms other baselines by a minimum of 5\% in F1 score. We obtain distinctively better results in 1-shot GFSID setup with gains of upto 20\% over the baselines. This shows that the proposed method is highly effective than other baselines in a challenging yet more practical setup. In DNNC, a pretrained entailment model is utilized to learn whether a pair of utterances are similar to each other or not. We conjecture that a similarity between  pair of utterances is different from an entailment task. Two sentences belonging to same class may not represent a premise and hypothesis pair, as hypothesis needs to entail from premise. Thus, we believe that entailment-pretrained model is not utilized efficiently in learning a few-shot model in DNNC. On the other hand, in NLI-FSL, the class-label name entails the utterance which provides a natural choice of hypothesis for utterance as a premise. Error analysis on the predictions of few-shot intents provides evidence in support of above mentioned arguments. For example, the intent of an utterance \textit{`listen newly added song'} is predicted to be `add to playlist' instead of `play music' by DNNC due to a high similarity with one of the nearest neighbor utterance \textit{`add new song'}. On the other hand, NLI-FSL correctly identify that `listen newly added song' \textit{entails} `play music'.

Another observation from results show that baseline methods have a bigger drop in performance going from a 5-shot setup to a more challenging 1-shot setup, when compared to results obtained in NLI-FSL method. For example, the drop in performance is 11.83\% for DNNC and 5.67\% for NLI-FSL in 1-shot FSID on SNIPS dataset (similar trend can be observed for other datasets). The observation shows that baseline methods struggles highly in a more challenging 1-shot settings, thus producing a non-competitive results compared to NLI-FSL. Additionally, the method have higher margins of improvement over baselines in GFSID settings compared to FSID. However, the overall F1-scores are lower in GFSID setup as compared to FSID. This demonstrates that predicting the intents in a joint label space of seen and unseen classes is a challenging task. Baseline methods comprising of prototypical networks, matching and attention based methods and discriminative nearest neighbour classification are ineffective in addressing the challenge of generalized few-shot detection.

Further analysis shows that NLI-FSL method is particularly effective for large label space datasets (BANKING77 and CLINC150), outperforming the most competitive baseline of DNNC by significant gains. In order to study the performance trends with variation in the size of label space, we evaluate our method against ProtoBERT, SMAN and DNNC by gradually increasing the size of novel label spaces. We consider CLINC150  dataset since it has the highest number of labels (150) among the datasets we have chosen. 
We vary the number of novel classes from 5 to 50 (10\% to 100\% of novel classes) for CLINC150 dataset. 
We present the average macro-F1 score over five runs of the experiment in a 5-shot FSID setting in Figure \ref{fig:inc_labels}. As expected, the performance of all the methods decreases as the number of novel classes increases. However, we can see that our method not only outperforms other baselines in every step of increment in size of label space, but is also more robust (lesser drop in results) with gradual addition of novel classes. Figure \ref{fig:inc_labels} shows that the results of SMAN and DNNC method on CLINC150 are closer to the NLI-FSL method when the number of novel (unseen) classes is lower but the performance gap increases on further addition of novel classes. 
Overall, from our results, we can say that the NLI-FSL method produces consistently better results compared to the prototype, semantic matching and discriminative nearest neighbor classifier based baselines as more novel classes are added. 
NLI-FSL, though, is compute intensive method due to its requirement to generate utterance-label pairs for all labels corresponding to the task.

\section{Conclusion}
We propose an effective method that formulates the task of few-shot intent detection to NLI based entailment task. We show that our method outperforms comparative works through extensive experiments on benchmark intent datasets on few-shot (FSID) and generalized few-shot (GFSID) setups.
Our method performs especially well under large label spaces in both 1-shot and 5-shot settings and highlights the limited performance of competitive approaches in challenging settings.
In future, we would like to 
extend the approach to utilize the multilingual language models for cross-lingual few-shot intent detection. 

\bibliographystyle{IEEEtran}

\bibliography{mybib}

\begin{thebibliography}{10}
\providecommand{\url}[1]{#1}
\csname url@samestyle\endcsname
\providecommand{\newblock}{\relax}
\providecommand{\bibinfo}[2]{#2}
\providecommand{\BIBentrySTDinterwordspacing}{\spaceskip=0pt\relax}
\providecommand{\BIBentryALTinterwordstretchfactor}{4}
\providecommand{\BIBentryALTinterwordspacing}{\spaceskip=\fontdimen2\font plus
\BIBentryALTinterwordstretchfactor\fontdimen3\font minus
  \fontdimen4\font\relax}
\providecommand{\BIBforeignlanguage}[2]{{%
\expandafter\ifx\csname l@#1\endcsname\relax
\typeout{** WARNING: IEEEtran.bst: No hyphenation pattern has been}%
\typeout{** loaded for the language `#1'. Using the pattern for}%
\typeout{** the default language instead.}%
\else
\language=\csname l@#1\endcsname
\fi
#2}}
\providecommand{\BIBdecl}{\relax}
\BIBdecl

\bibitem{DBLP:conf/cvpr/MillerMV00}
\BIBentryALTinterwordspacing
E.~G. Miller, N.~E. Matsakis, and P.~A. Viola, ``Learning from one example
  through shared densities on transforms,'' in \emph{2000 Conference on
  Computer Vision and Pattern Recognition {(CVPR} 2000), 13-15 June 2000,
  Hilton Head, SC, {USA}}.\hskip 1em plus 0.5em minus 0.4em\relax {IEEE}
  Computer Society, 2000, pp. 1464--1471. [Online]. Available:
  \url{https://doi.org/10.1109/CVPR.2000.855856}
\BIBentrySTDinterwordspacing

\bibitem{DBLP:journals/corr/abs-2004-01881}
\BIBentryALTinterwordspacing
C.~Xia, C.~Zhang, H.~Nguyen, J.~Zhang, and P.~S. Yu, ``{CG-BERT:} conditional
  text generation with {BERT} for generalized few-shot intent detection,''
  \emph{CoRR}, vol. abs/2004.01881, 2020. [Online]. Available:
  \url{https://arxiv.org/abs/2004.01881}
\BIBentrySTDinterwordspacing

\bibitem{DBLP:conf/nips/SnellSZ17}
J.~Snell, K.~Swersky, and R.~S. Zemel, ``Prototypical networks for few-shot
  learning,'' in \emph{Advances in Neural Information Processing Systems 30:
  Annual Conference on Neural Information Processing Systems 2017, December
  4-9, 2017, Long Beach, CA, {USA}}, I.~Guyon, U.~von Luxburg, S.~Bengio, H.~M.
  Wallach, R.~Fergus, S.~V.~N. Vishwanathan, and R.~Garnett, Eds., 2017, pp.
  4077--4087.

\bibitem{xu2021semantic}
W.~Xu, P.~Zhou, C.~You, and Y.~Zou, ``Semantic transportation prototypical
  network for few-shot intent detection,'' \emph{Proc. Interspeech 2021}, pp.
  251--255, 2021.

\bibitem{sun2019hierarchical}
S.~Sun, Q.~Sun, K.~Zhou, and T.~Lv, ``Hierarchical attention prototypical
  networks for few-shot text classification,'' in \emph{Proceedings of the 2019
  Conference on Empirical Methods in Natural Language Processing and the 9th
  International Joint Conference on Natural Language Processing
  (EMNLP-IJCNLP)}, 2019, pp. 476--485.

\bibitem{nguyen-etal-2020-dynamic}
\BIBentryALTinterwordspacing
H.~Nguyen, C.~Zhang, C.~Xia, and P.~Yu, ``Dynamic semantic matching and
  aggregation network for few-shot intent detection,'' in \emph{Findings of the
  Association for Computational Linguistics: EMNLP 2020}.\hskip 1em plus 0.5em
  minus 0.4em\relax Online: Association for Computational Linguistics, Nov.
  2020, pp. 1209--1218. [Online]. Available:
  \url{https://aclanthology.org/2020.findings-emnlp.108}
\BIBentrySTDinterwordspacing

\bibitem{ye-ling-2019-multi}
\BIBentryALTinterwordspacing
Z.-X. Ye and Z.-H. Ling, ``Multi-level matching and aggregation network for
  few-shot relation classification,'' in \emph{Proceedings of the 57th Annual
  Meeting of the Association for Computational Linguistics}.\hskip 1em plus
  0.5em minus 0.4em\relax Florence, Italy: Association for Computational
  Linguistics, Jul. 2019, pp. 2872--2881. [Online]. Available:
  \url{https://aclanthology.org/P19-1277}
\BIBentrySTDinterwordspacing

\bibitem{finn2017model}
C.~Finn, P.~Abbeel, and S.~Levine, ``Model-agnostic meta-learning for fast
  adaptation of deep networks,'' in \emph{International Conference on Machine
  Learning}.\hskip 1em plus 0.5em minus 0.4em\relax PMLR, 2017, pp. 1126--1135.

\bibitem{DBLP:conf/slt/0001KDZA21}
\BIBentryALTinterwordspacing
S.~Li, J.~Krone, S.~Dong, Y.~Zhang, and Y.~Al{-}onaizan, ``Meta learning to
  classify intent and slot labels with noisy few shot examples,'' in
  \emph{{IEEE} Spoken Language Technology Workshop, {SLT} 2021, Shenzhen,
  China, January 19-22, 2021}.\hskip 1em plus 0.5em minus 0.4em\relax {IEEE},
  2021, pp. 1004--1011. [Online]. Available:
  \url{https://doi.org/10.1109/SLT48900.2021.9383489}
\BIBentrySTDinterwordspacing

\bibitem{xia-etal-2020-composed}
\BIBentryALTinterwordspacing
C.~Xia, C.~Xiong, P.~Yu, and R.~Socher, ``Composed variational natural language
  generation for few-shot intents,'' in \emph{Findings of the Association for
  Computational Linguistics: EMNLP 2020}.\hskip 1em plus 0.5em minus
  0.4em\relax Online: Association for Computational Linguistics, Nov. 2020, pp.
  3379--3388. [Online]. Available:
  \url{https://aclanthology.org/2020.findings-emnlp.303}
\BIBentrySTDinterwordspacing

\bibitem{10.1145/3404835.3462995}
\BIBentryALTinterwordspacing
C.~Xia, C.~Xiong, and P.~Yu, \emph{Pseudo Siamese Network for Few-Shot Intent
  Generation}.\hskip 1em plus 0.5em minus 0.4em\relax New York, NY, USA:
  Association for Computing Machinery, 2021, p. 2005–2009. [Online].
  Available: \url{https://doi.org/10.1145/3404835.3462995}
\BIBentrySTDinterwordspacing

\bibitem{bowman-etal-2015-large}
\BIBentryALTinterwordspacing
S.~R. Bowman, G.~Angeli, C.~Potts, and C.~D. Manning, ``A large annotated
  corpus for learning natural language inference,'' in \emph{Proceedings of the
  2015 Conference on Empirical Methods in Natural Language Processing}.\hskip
  1em plus 0.5em minus 0.4em\relax Lisbon, Portugal: Association for
  Computational Linguistics, Sep. 2015, pp. 632--642. [Online]. Available:
  \url{https://aclanthology.org/D15-1075}
\BIBentrySTDinterwordspacing

\bibitem{DBLP:conf/emnlp/YinRRSX20}
\BIBentryALTinterwordspacing
W.~Yin, N.~F. Rajani, D.~R. Radev, R.~Socher, and C.~Xiong, ``Universal natural
  language processing with limited annotations: Try few-shot textual entailment
  as a start,'' in \emph{Proceedings of the 2020 Conference on Empirical
  Methods in Natural Language Processing, {EMNLP} 2020, Online, November 16-20,
  2020}, B.~Webber, T.~Cohn, Y.~He, and Y.~Liu, Eds.\hskip 1em plus 0.5em minus
  0.4em\relax Association for Computational Linguistics, 2020, pp. 8229--8239.
  [Online]. Available: \url{https://doi.org/10.18653/v1/2020.emnlp-main.660}
\BIBentrySTDinterwordspacing

\bibitem{sainz-etal-2021-label}
\BIBentryALTinterwordspacing
O.~Sainz, O.~Lopez~de Lacalle, G.~Labaka, A.~Barrena, and E.~Agirre, ``Label
  verbalization and entailment for effective zero and few-shot relation
  extraction,'' in \emph{Proceedings of the 2021 Conference on Empirical
  Methods in Natural Language Processing}.\hskip 1em plus 0.5em minus
  0.4em\relax Online and Punta Cana, Dominican Republic: Association for
  Computational Linguistics, Nov. 2021, pp. 1199--1212. [Online]. Available:
  \url{https://aclanthology.org/2021.emnlp-main.92}
\BIBentrySTDinterwordspacing

\bibitem{zhang-etal-2020-discriminative}
\BIBentryALTinterwordspacing
J.~Zhang, K.~Hashimoto, W.~Liu, C.-S. Wu, Y.~Wan, P.~Yu, R.~Socher, and
  C.~Xiong, ``Discriminative nearest neighbor few-shot intent detection by
  transferring natural language inference,'' in \emph{Proceedings of the 2020
  Conference on Empirical Methods in Natural Language Processing
  (EMNLP)}.\hskip 1em plus 0.5em minus 0.4em\relax Online: Association for
  Computational Linguistics, Nov. 2020, pp. 5064--5082. [Online]. Available:
  \url{https://aclanthology.org/2020.emnlp-main.411}
\BIBentrySTDinterwordspacing

\bibitem{vinyals2016matching}
O.~Vinyals, C.~Blundell, T.~Lillicrap, D.~Wierstra \emph{et~al.}, ``Matching
  networks for one shot learning,'' \emph{Advances in neural information
  processing systems}, vol.~29, pp. 3630--3638, 2016.

\bibitem{laenen2021on}
\BIBentryALTinterwordspacing
S.~Laenen and L.~Bertinetto, ``On episodes, prototypical networks, and few-shot
  learning,'' in \emph{Advances in Neural Information Processing Systems},
  A.~Beygelzimer, Y.~Dauphin, P.~Liang, and J.~W. Vaughan, Eds., 2021.
  [Online]. Available: \url{https://openreview.net/forum?id=bJaZ8leI0QJ}
\BIBentrySTDinterwordspacing

\bibitem{devlin-etal-2019-bert}
\BIBentryALTinterwordspacing
J.~Devlin, M.-W. Chang, K.~Lee, and K.~Toutanova, ``{BERT}: Pre-training of
  deep bidirectional transformers for language understanding,'' in
  \emph{Proceedings of the 2019 Conference of the North {A}merican Chapter of
  the Association for Computational Linguistics: Human Language Technologies,
  Volume 1 (Long and Short Papers)}.\hskip 1em plus 0.5em minus 0.4em\relax
  Minneapolis, Minnesota: Association for Computational Linguistics, Jun. 2019,
  pp. 4171--4186. [Online]. Available: \url{https://aclanthology.org/N19-1423}
\BIBentrySTDinterwordspacing

\bibitem{Coucke2018SnipsVP}
A.~Coucke, A.~Saade, A.~Ball, T.~Bluche, A.~Caulier, D.~Leroy, C.~Doumouro,
  T.~Gisselbrecht, F.~Caltagirone, T.~Lavril, M.~Primet, and J.~Dureau, ``Snips
  voice platform: an embedded spoken language understanding system for
  private-by-design voice interfaces,'' \emph{ArXiv}, vol. abs/1805.10190,
  2018.

\bibitem{hemphill-etal-1990-atis}
\BIBentryALTinterwordspacing
C.~T. Hemphill, J.~J. Godfrey, and G.~R. Doddington, ``The {ATIS} spoken
  language systems pilot corpus,'' in \emph{Speech and Natural Language:
  Proceedings of a Workshop Held at Hidden Valley, {P}ennsylvania, June
  24-27,1990}, 1990. [Online]. Available:
  \url{https://aclanthology.org/H90-1021}
\BIBentrySTDinterwordspacing

\bibitem{DBLP:journals/corr/abs-2106-04564}
\BIBentryALTinterwordspacing
J.~Zhang, K.~Hashimoto, Y.~Wan, Y.~Liu, C.~Xiong, and P.~S. Yu, ``Are
  pretrained transformers robust in intent classification? {A} missing
  ingredient in evaluation of out-of-scope intent detection,'' \emph{CoRR},
  vol. abs/2106.04564, 2021. [Online]. Available:
  \url{https://arxiv.org/abs/2106.04564}
\BIBentrySTDinterwordspacing

\bibitem{larson-etal-2019-evaluation}
\BIBentryALTinterwordspacing
S.~Larson, A.~Mahendran, J.~J. Peper, C.~Clarke, A.~Lee, P.~Hill, J.~K.
  Kummerfeld, K.~Leach, M.~A. Laurenzano, L.~Tang, and J.~Mars, ``An evaluation
  dataset for intent classification and out-of-scope prediction,'' in
  \emph{Proceedings of the 2019 Conference on Empirical Methods in Natural
  Language Processing and the 9th International Joint Conference on Natural
  Language Processing (EMNLP-IJCNLP)}.\hskip 1em plus 0.5em minus 0.4em\relax
  Hong Kong, China: Association for Computational Linguistics, Nov. 2019, pp.
  1311--1316. [Online]. Available: \url{https://aclanthology.org/D19-1131}
\BIBentrySTDinterwordspacing

\bibitem{gao2019hybrid}
T.~Gao, X.~Han, Z.~Liu, and M.~Sun, ``Hybrid attention-based prototypical
  networks for noisy few-shot relation classification,'' in \emph{Proceedings
  of the AAAI Conference on Artificial Intelligence}, vol.~33, no.~01, 2019,
  pp. 6407--6414.

\end{thebibliography}


\end{document}